\pdfoutput=1

\documentclass[11pt]{article}

\usepackage{emnlp2021}

\usepackage{times}
\usepackage{latexsym}

\usepackage[T1]{fontenc}

\usepackage[utf8]{inputenc}

\usepackage{microtype}

\usepackage{tabularx}
\usepackage{booktabs}
\usepackage{graphicx}
\usepackage{colortbl}
\usepackage{soul}
\usepackage{hyperref}
\usepackage{amsmath}
\usepackage{amssymb}
\usepackage{multirow} 
\usepackage{balance}
\usepackage{inconsolata}

\def\rot{\rotatebox}

\definecolor{darkspringgreen}{rgb}{0.09, 0.45, 0.27}
\definecolor{darkred}{rgb}{0.55, 0.0, 0.0}

\definecolor{alizarin}{rgb}{0.82, 0.1, 0.26}
\definecolor{ao(english)}{rgb}{0.0, 0.5, 0.0}
\definecolor{cadmiumgreen}{rgb}{0.0, 0.42, 0.24}

\newcommand{\kb}{KB}

\newcommand{\lm}{LM}

\newcommand{\xPos}{x^{+}}
\newcommand{\triple}{(X_h, r, X_t)}
\newcommand{\xNeg}{x^{-}}
\newcommand{\xCorr}{\tilde{x}}

\newcommand{\loss}{\mathcal{L}}
\newcommand{\grad}{\tilde{M}}
\newcommand{\CLS}{\texttt{[CLS]}}
\newcommand{\SEP}{\texttt{[SEP]}}

\newcommand{\cn}{ConceptNet}

\newcommand{\rel}[1]{\textsc{#1}}

\newcommand{\negater}{\textsc{NegatER}}

\newcommand{\threshold}{\theta_r}
\newcommand{\negaterThresh}{\textsc{NegatER}-$\threshold$}
\newcommand{\negaterGrad}{\textsc{NegatER}-$\nabla$}
\newcommand{\proxy}{\tilde{f_M}}
\newcommand{\uniform}{\textsc{Uniform}}
\newcommand{\slots}{\textsc{Slots}}
\newcommand{\antonyms}{\textsc{Antonyms}}

\newcommand{\sans}{\textsc{SANS}}
\newcommand{\selfadv}{\textsc{RotatE-SA}}
\newcommand{\comet}{\textsc{COMeT}}

\newcommand{\grammar}{\textbf{R1}}
\newcommand{\consistency}{\textbf{R2}}
\newcommand{\contradiction}{\textbf{R3}}

\newcommand*\pct{\scalebox{.9}{\%}}

\title{\negater: Unsupervised Discovery of \\ Negatives in Commonsense Knowledge Bases}

\author{
  Tara Safavi, Jing Zhu, Danai Koutra \\
  University of Michigan, Ann Arbor \\
  \texttt{\{tsafavi,jingzhuu,dkoutra\}@umich.edu}
}

\date{}

\begin{document}
\maketitle
\begin{abstract}
Codifying commonsense knowledge in machines is a longstanding goal of artificial intelligence.
Recently, much progress toward this goal has been made with automatic knowledge base (KB) construction techniques.
However, such techniques focus primarily on the acquisition of positive (true) KB statements, even though \emph{negative} (false) statements are often also important for discriminative reasoning over commonsense KBs. 
As a first step toward the latter, this paper proposes NegatER, a framework that ranks potential negatives in commonsense KBs using a contextual language model (LM). 
Importantly, as most KBs do not contain negatives, NegatER relies only on the \emph{positive} knowledge in the LM and does not require ground-truth negative examples. 
Experiments demonstrate that, compared to multiple contrastive data augmentation approaches, NegatER yields negatives that are more grammatical, coherent, and informative---leading to statistically significant accuracy improvements in a challenging KB completion task and confirming that the positive knowledge in LMs can be ``re-purposed'' to generate negative knowledge.
\end{abstract}

\section{Introduction}
\label{sec:intro}
Endowing machines with \emph{commonsense}, which is knowledge that members of a culture usually agree upon but do not express explicitly, is a major but elusive goal of artificial intelligence~\cite{minsky1974framework,davis1993knowledge,liu-singh-2004-conceptnet,davis2015commonsense}. 
One way to capture such knowledge is with curated commonsense knowledge bases (\kb{}s), which contain semi-structured statements of ``everyday'' human knowledge. 
As such \kb{}s are increasingly being used to augment the capabilities of intelligent agents~\cite{hwang-etal-2021-comet-atomic}, 
automatically expanding their scope has become  crucial~\cite{li-etal-2016-commonsense,davison-etal-2019-commonsense,bosselut-etal-2019-comet,malaviya-etal-2020-commonsense}.

\begin{table}[t!]
\centering
\caption{
    Out-of-\kb{} statements are less meaningful as negative examples when sampled at random versus ranked with our \negater{} framework.
    The random examples are taken from the test split of the \cn{} benchmark introduced by~\citet{li-etal-2016-commonsense}. 
}
\label{table:examples}
\resizebox{\columnwidth}{!}{
    \begin{tabular}{ l l }
    \toprule
    Method & Negative statement \\ 
    \toprule 
    \multirow{3}{*}{Random sampling} & (``tickle'', \rel{HasSubevent}, ``supermarket'') \\
    & (``lawn mower'', \rel{AtLocation}, ``pantry'') \\ 
    & (``closet'', \rel{UsedFor}, ``play baseball'') \\ 
    \midrule 
    \multirow{3}{*}{\negater{} ranking} & (``ride horse'', \rel{HasSubevent}, ``pedal'') \\
    & (``zoo keeper'', \rel{AtLocation}, ``jungle'') \\
    & (``air ticket'', \rel{UsedFor}, ``get onto trolley'') \\ 
    \bottomrule
\end{tabular}
}
\end{table}

Previous research in this direction focuses primarily on the acquisition of positive knowledge, or that which is true about the world. 
However, understanding what is true about the world often also requires gathering and reasoning over explicitly \emph{untrue} information. 
Humans routinely rely on \textbf{negative knowledge}---that is, what ``not to do'' or what ``not to believe''---in order to increase certainty in decision-making and avoid mistakes and accidents~\cite{minsky1994negative}. 
Similarly, discriminative models that operate over structured knowledge from \kb{}s often require explicit negative examples in order to learn good decision boundaries~\cite{sun-etal-2019-rotate,ahrabian-etal-2020-structure,ma-etal-2021-knowledge-driven}. 

The main challenge with machine acquisition of structured negative knowledge, commonsense or otherwise, is that most \kb{}s do not contain negatives at all~\cite{safavi-koutra-2020-codex,arnaout-etal-2020-enriching}. 
Therefore, for \kb-related tasks that require both positive and negative statements, negatives must either be gathered via human annotation, or else generated ad-hoc.
Both of these approaches entail distinct challenges. 
On one hand, human annotation of negatives can be cost-prohibitive at scale.
On the other, automatic negative generation without good training examples can lead to uninformative, even nonsensical statements (Table~\ref{table:examples}), because the prevailing approach is to randomly sample negatives from the large space of all out-of-\kb{} statements~\cite{li-etal-2016-commonsense}. 

To strike a balance between expert annotation, which is costly but accurate, and random sampling, which is efficient but inaccurate, we propose \textbf{\negater}, a framework for unsupervised discovery of \textbf{Negat}ive Commonsense Knowledge in \textbf{E}ntity and \textbf{R}elation form.
Rather than randomly sampling from the space of all out-of-\kb{} statements to obtain negatives, 
\negater{} \emph{ranks} a selection of these statements such that higher-ranking statements are ``more likely'' to be negative.
Ranking is done with a fine-tuned contextual language model (\lm), building upon studies showing that \lm{}s can be trained to acquire a degree of commonsense ``knowledge''~\cite{petroni-etal-2019-language}. 

Importantly, because we do not assume the presence of gold negative examples for training the \lm{}, we devise techniques that make use of \emph{positive \kb{} statements only}.
This distinguishes \negater{} from supervised generative commonsense \kb{} construction techniques that require abundant gold examples, 
usually obtained via human annotation, for  fine-tuning~\cite{bosselut-etal-2019-comet,hwang-etal-2021-comet-atomic,jiang-etal-2021-im}.
Our realistic assumption means that we do not have any explicit examples of true negatives and therefore cannot guarantee a minimum true negative rate; indeed,  obtaining true negatives in \kb{}s is a hard problem in general~\cite{arnaout-etal-2020-enriching}. 
However, we show in detailed experiments that \negater{} strikes a delicate balance between several factors that contribute to high-quality negative knowledge, including task-specific utility, coherence, and the true negative rate. 
Our contributions are as follows:
\begin{itemize}
    \setlength\itemsep{0.0001em}
    \item \textbf{Problem definition}: 
    We provide the first rigorous definition of negative knowledge in commonsense \kb{}s (\S~\ref{sec:problem-definition}), 
    which as far as we are aware has not been studied before.
    \item \textbf{Framework}: 
    We introduce \negater{} (\S~\ref{sec:method}), which ranks out-of-\kb{} potential negatives using a contextual language model (\lm). 
    As \kb{}s typically do not contain gold negatives, we devise an approach that relies only on the \lm{}'s \emph{positive} beliefs.
    Specifically, 
    \negater{} first fine-tunes the \lm{} to acquire high-quality positive knowledge, then ranks  potential negatives by how much they ``contradict'' the \lm{}'s positive knowledge, as measured by its classification scores or gradients. 
    \item \textbf{Evaluation}:
    In keeping with the novelty of the problem, we conduct multiple evaluations that address the fundamental  research questions of negative commonsense. 
    First, we measure the effectiveness of our \lm{} fine-tuning approach and the utility of \negater{}-generated negatives in \kb{} completion tasks (\S~\ref{eval:lm},~\ref{eval:extrinsic}). 
    Next, we study the intrinsic quality of the generated negatives (\S~\ref{eval:intrinsic}).
    When considering all such factors, \negater{} outperforms numerous competitive baselines.
    Most notably, training \kb{} completion models with highly-ranked negative examples from \negater{} results in statistically significant accuracy improvements of up to 1.90\pct{}. 
    Code and data are available at \url{https://github.com/tsafavi/NegatER}. 
\end{itemize}

\section{Problem definition}
\label{sec:problem-definition}
As the problem of negative knowledge has not yet been addressed in the commonsense \kb{} completion literature, we begin by defining meaningful negatives in commonsense \kb{}s. 

\subsection{Positive knowledge}
\label{problem:kbs}

\begin{figure*}
    \centering
    \includegraphics[width=0.99\textwidth]{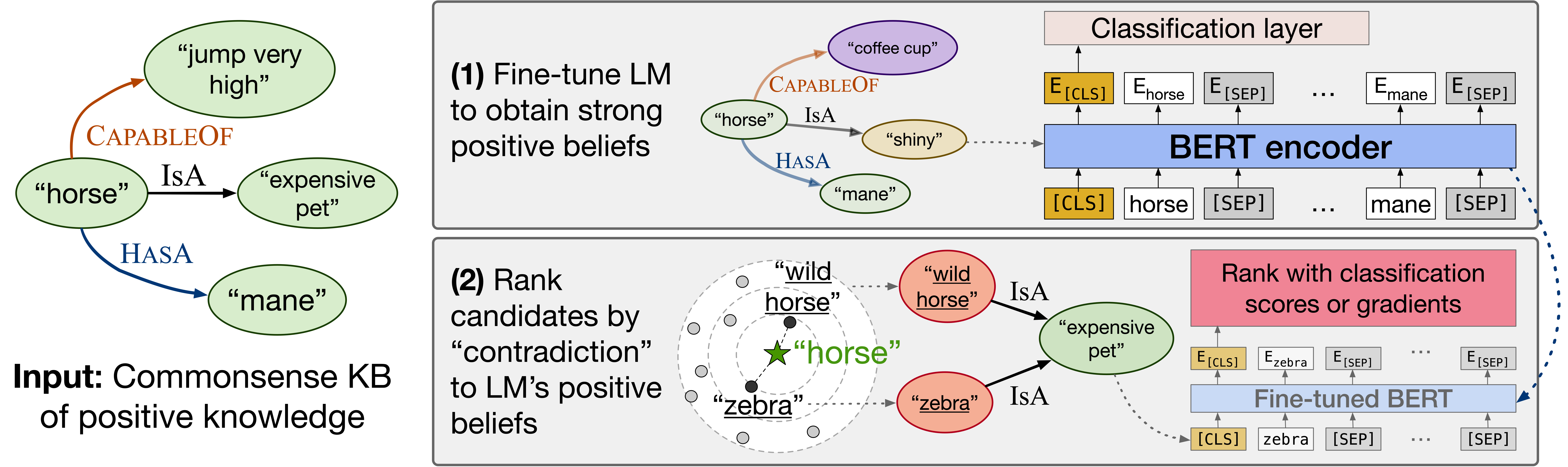}  
    \caption{
    \negater{} consists of two steps: 
    \textbf{(1)}~Fine-tuning an \lm{} on the input \kb{} to obtain strong positive beliefs; and \textbf{(2)}~Feeding a set of out-of-\kb{} candidate statements to the fine-tuned \lm{} and ranking them by the \lm{}'s classification scores or gradients. 
    Here, the \kb{} is a fragment of \cn{}~\cite{speer-havasi-2012-representing}. 
    }
    \label{fig:negative-generation}
\end{figure*}

A commonsense knowledge base (\kb) consists of triples $\{ \xPos \} = \{ \triple^{+} \}$, where the superscript denotes that all in-\kb{} triples are assumed to be positive or true. 
In each triple, the first and third terms are head and tail entities in the form of phrases $X_h = [w_1, \hdots, w_{h}]$ and $X_t = [w_1, \hdots, w_{t}]$ drawn from a potentially infinite vocabulary. 
The relation types $r$ are symbolic and drawn from a finite dictionary $R$. 
Figure~\ref{fig:negative-generation} provides examples of positive statements from the \cn{} \kb~\cite{speer-havasi-2012-representing}, e.g., ($X_h$=``horse'', $r$=\rel{IsA}, $X_t$=``expensive pet''). 

\subsection{Negative knowledge}
\label{problem:negative-knowledge}

We denote a negative triple as $\xNeg \not\in \{ \xPos \}$.
As the space of negatives is evidently much larger than the space of positives, we define negative knowledge to exclude \emph{trivial negatives}, for example simple negations or nonsensical statements. 
Specifically, drawing from the literature on procedural negative expertise in humans~\cite{minsky1994negative,gartmeier-etal-2008-negative}, we define negative knowledge as 
\ul{nonviable or explicitly false knowledge that is heuristically valuable with respect to a given task, goal, or decision}.
In the context of \kb{}s, we devise three requirements that, combined, satisfy this definition: 
\begin{itemize}
    \setlength{\itemsep}{0.0001em}
    \item [\grammar{}] Negative knowledge must resemble positive knowledge in structure.
    This means that negative statements should \textbf{obey the grammatical rules} (parts of speech) of their relation types. 
    \item [\consistency{}] The head and tail phrases must be \textbf{thematically or topically consistent}. 
    For example, given the head phrase $X_h$=``make coffee,''
    a consistent tail phrase is one that is thematically related but still nonviable with respect to the whole statement, for example (``make coffee'', \rel{HasSubevent}, ``buy tea''). 
    \item [\contradiction{}] 
    Negative knowledge must be informative for a given task, goal, or decision.
    We consider a statement as informative if, when taken as true, it is \textbf{counterproductive or contradictory} to the goal at hand,
    e.g., (``make coffee'', \rel{HasSubevent}, ``drop mug'').
\end{itemize}

\section{Framework}
\label{sec:method}
We propose the \negater{} framework to match our definition of negative knowledge. 
As shown in Figure~\ref{fig:negative-generation}, \negater{} consists of two steps:
First, a pretrained \lm{} is fine-tuned on a given commonsense \kb{} using a contrastive approach to acquire strong positive beliefs. 
Then, a set of grammatical (\grammar) and topically consistent (\consistency) out-of-\kb{} candidate statements are fed to the \lm{} and ranked by the degree to which they ``contradict'' the \lm's fine-tuned positive beliefs (\contradiction), such that the higher-ranking statements are more likely to be negative. 
We emphasize that ground-truth negative examples are \emph{not required} at any point, which means that we trade off some accuracy (i.e., the true negative rate) for cost efficiency (i.e., the cost of gathering ground-truth negative examples for training via expert annotation). 

We describe each step in detail in the remainder of this section. 

\subsection{Fine-tuning for positive knowledge}
\label{method:lm}

The first step of \negater{} is to minimally fine-tune a language model on a given commonsense \kb{} using contrastive learning (step 1, Figure~\ref{fig:negative-generation}), such that it acquires strong positive beliefs. 
We focus on encoder-only BERT-based models~\cite{devlin-etal-2019-bert,liu-etal-2019-roberta}, as we will ultimately use their fine-tuned encodings to represent triples. 

\paragraph{\lm{} input and output}
We input a \kb{} triple $\triple$ to the \lm{} by concatenating BERT's special \CLS{} token with a linearized version of the triple, delineating the head tokens $X_h$, the relation $r$, and the tail tokens $X_t$ with BERT's special \SEP{} token.
At the output of the encoder, we apply a semantic-level pooling operation (e.g., any of those proposed by~\citet{reimers-gurevych-2019-sentence}) to obtain a single contextual representation of the triple, and pass it through a classification layer $W \in \mathbb{R}^{H}$, where $H$ is the hidden layer dimension. 

\paragraph{Supervision strategy}
Since the goal of fine-tuning is to endow the \lm{} with strong positive beliefs, we use a common contrastive data augmentation technique for positive \kb{} triple classification~\cite{li-etal-2016-commonsense,malaviya-etal-2020-commonsense}.
Specifically, for each positive $\xPos$, we construct a contrastive corrupted version where the head, relation, or tail has been replaced by a random phrase or relation from the \kb{}.
We minimize binary cross-entropy loss between the positive training examples and their corrupted counterparts. 
We learn a decision threshold $\threshold$ per relation $r$ on the validation set to maximize validation accuracy, such that triples of relation $r$ scored above $\threshold$ are classified as positive. 

\subsection{Ranking out-of-\kb{} statements}
\label{method:corruptions}

Now that we have an \lm{} fine-tuned to a given commonsense \kb{}, we 
feed a set of out-of-\kb{} candidate statements to the \lm{} in the same format as was used during fine-tuning, and rank them by the degree to which they ``contradict'' the \lm{}'s positive beliefs (step 2, Figure~\ref{fig:negative-generation}).

\paragraph{Out-of-\kb{} candidate generation}
To gather out-of-\kb{} candidate statements, we use a dense $k$-nearest-neighbors retrieval approach.
The idea here is that the set of \emph{all} out-of-\kb{} statements is extremely large and most such statements are not likely to be meaningful, so we narrow the candidates down to a smaller set that is more likely to be grammatical (\grammar) and consistent (\consistency).

For each positive triple $\xPos = \triple^{+}$, we retrieve the $k$ nearest-neighbor phrases to head phrase $X_h$ using a maximum inner product search~\cite{johnson2019billion} over pre-computed embeddings of the \kb{}'s entity phrases.
While any choice of embedding and distance measure may be used, we use Euclidean distance between the \CLS{} embeddings output by a separate pretrained BERT model for its empirical good performance. 
We then replace $X_h$ in the head slot of the original positive $\xPos$ by each of its neighbors $\tilde{X}_h$ in turn, yielding a set of candidates 
$$
\{\xCorr\}_{i=1}^{k}, \,\,\, \xCorr = (\tilde{X}_h, r, X_t).
$$ 
We discard any candidates that already appear in the \kb{} and repeat this process for the tail phrase $X_t$, yielding up to $2k$ candidates $\xCorr$ per positive $\xPos$.
We also filter the candidates to only those for which the retrieved head (tail) phrase $\tilde{X}_h$ ($\tilde{X}_t$) appears in the head (tail) slot of relation $r$ in the \kb{}. 

\paragraph{Meeting R1, R2, and R3}
Our filtering process discards candidate triples whose head/tail entities have not been observed to co-occur with relation $r$, which preserves the grammar (\grammar) of the relation. 
Notice that by retrieving the nearest neighbors of each head and tail phrase by semantic similarity, we also preserve the topical consistency (\consistency) of the original positive statements.

Finally, to meet requirement \contradiction{}, we rank the remaining out-of-\kb{} candidates by the degree to which they ``contradict'' the positive beliefs of the fine-tuned \lm{}. 
These ranked statements can be then taken in order of rank descending as input to any discriminative \kb{} reasoning task requiring negative examples, with the exact number of negatives being determined by the practitioner and application. 
We propose two independent ranking strategies: 

\subsubsection{\negaterThresh: Ranking with scores}
\label{method:negater-thresh}

Our first approach, \negaterThresh{}, relies on the decision thresholds $\threshold{}$ set during the validation stage of fine-tuning.
We feed the candidates $\xCorr$ to the \lm{} and take only those that the \lm{} classifies below the respective decision threshold $\threshold$. 
Per relation $r$, the candidates are ranked descending by their scores at the output of the classification layer, such that the higher-ranking candidates look more plausible---that is, ``almost positive''---to the \lm. 

\subsubsection{\negaterGrad{}: Ranking with gradients}
\label{method:negater-grad}

The premise of our second approach, \negaterGrad{}, is that the candidates that most ``surprise'' the \lm{} \emph{when labeled as true} are the most likely to be negative, because they most directly contradict what the \lm{} has observed during fine-tuning.

We quantify ``surprisal'' with the \lm{}'s gradients.
Let $\loss(\xCorr; \tilde{y})$ be the binary cross-entropy loss evaluated on candidate $\xCorr$ given a corresponding label $\tilde{y} \in \{-1, 1\}$. 
We feed each $\xCorr$ to the \lm{} and compute the magnitude of the gradient of $\loss$ with respect to the \lm's parameters $\Theta$, given a positive labeling of $\xCorr$: 
\begin{align}
    \grad{} = \left\| 
    \frac{\partial \mathcal{L}(\xCorr; \tilde{y} = 1)}{\partial \Theta } \right\|,
\end{align} 
and rank candidates in descending order of gradient magnitude $\grad{}$. 
Here, $\grad{}$ signifies the amount to which the \lm's fine-tuned beliefs would need to be updated to incorporate this candidate as positive. 
Therefore, the higher the $\grad{}$, the more directly $\xCorr$ contradicts or negates the \lm's positive beliefs.


\paragraph{Faster computation}
Because \negaterGrad{} requires a full forward and backward pass for each candidate $\xCorr$, it can be costly for a large number $N$ of  candidates.
We therefore propose a simple (optional) trick to speed up computation. 
We first compute $\grad{}$ for an initial sample of $n \ll N$ candidates. 
We then use the contextual representations of these $n$ candidates and their gradient magnitudes $\grad{}$ as training features and targets, respectively, to learn a regression function $\proxy{}: \mathbb{R}^H \rightarrow \mathbb{R}$. 
Finally, we substitute the \lm{}'s fine-tuning layer with $\proxy{}$, allowing us to skip the backward pass and feed batches of candidates $\xCorr$ to the \lm{} in forward passes.  
In our experiments, we will show that this approach is an effective and efficient alternative to full-gradient computation.

\paragraph{Gradients versus losses}
On the surface, it might seem that \negaterGrad{} could be made more efficient by ranking examples descending by their \emph{losses}, instead of gradients. 
However, notice that the binary cross-entropy loss $\loss(\xCorr; \tilde{y} = 1)$ is low for candidates $\xCorr$ that receives high scores from the \lm{}, and high for candidates that receive low scores.
Due to the contrastive approach that we used for fine-tuning, candidates with the lowest losses are mainly \emph{true} statements, and candidates with the highest losses are mainly \emph{nonsensical} statements.
Therefore, the losses do not directly correlate with how ``contradictory'' the candidate statements are.
By contrast, the gradient-based approach quantifies how much the \lm{} would need to change its beliefs to incorporate the new knowledge as positive, which more directly matches requirement \contradiction.


\section{Fine-tuning evaluation}
\label{eval:lm}

In this section, we evaluate the efficacy of the fine-tuning step of \negater{} (\S~\ref{method:lm}). 
In the following sections, we will evaluate the efficacy of the ranking step of \negater{} (\S~\ref{method:corruptions}) from quantitative and qualitative perspectives.

\subsection{Data}
The goal of this experiment is to evaluate whether our fine-tuning strategy from \S~\ref{method:lm} endows \lm{}s with sufficiently accurate positive knowledge.
For this, we use the English-language triple classification benchmark introduced by \citet{li-etal-2016-commonsense}, which consists of 100K/2400/2400 train/validation/test triples across 34 relations and 78,334 unique entity phrases from \cn{} 5~\cite{speer-havasi-2012-representing}. 
The evaluation metric is accuracy.
In the evaluation splits, which are balanced positive/negative 50/50, the negatives were constructed by swapping the head, relation, or tail of each positive $\xPos$ with that of another randomly sampled positive from the \kb. 
Note that while the test negatives were generated randomly and are therefore mostly nonsensical (Table~\ref{table:examples}), we use this benchmark because it mainly tests models' recognition of positive knowledge, which matches the goals of our fine-tuning procedure.
Ultimately, however, a more difficult dataset will be needed, which we will introduce in the next section.


\subsection{Methods}
We fine-tune BERT-\textsc{Base-Uncased}~\cite{devlin-etal-2019-bert} and RoBERTa-\textsc{Base}~\cite{liu-etal-2019-roberta}.
We compare our \lm{}s to all published results on the same evaluation splits of which we are aware.
Our baselines include both \kb{} embeddings~\cite{li-etal-2016-commonsense,jastrzebski-etal-2018-commonsense} and contextual \lm{}s~\cite{davison-etal-2019-commonsense,shen-etal-2020-exploiting}.  
Appendix~\ref{appendix:fine-tuning} provides implementation details. 

\subsection{Results}
The results in Table~\ref{table:tc} confirm the effectiveness of our fine-tuning approach, as our BERT reaches state-of-the-art accuracy on \cn{}. 
It even outperforms KG-BERT$_{\textrm{GLM(RoBERTa-\textsc{Large})}}$~\cite{shen-etal-2020-exploiting}, which requires an entity linking step during preprocessing and uses a RoBERTa-\textsc{Large} model pretrained with several extra tasks. 
In fact, we suspect that our fine-tuned BERT has saturated this benchmark, as it slightly exceeds the human accuracy estimate provided by~\citet{li-etal-2016-commonsense}. 

\begin{table}[t!]
	\centering
	\caption{
	Our fine-tuned BERT reaches state-of-the-art accuracy on the \cn{} benchmark from~\cite{li-etal-2016-commonsense}. 
	Baseline results are reported directly from the referenced papers. 
	}
	\label{table:tc}
    \resizebox{0.99\columnwidth}{!}{
    \begin{tabular}{ l l  }
        \toprule
        & Acc. \\ 
        \toprule 
        Bilinear AVG~\cite{li-etal-2016-commonsense} &  91.70 \\ 
        DNN AVG~\cite{li-etal-2016-commonsense} &  92.00 \\ 
        DNN LSTM~\cite{li-etal-2016-commonsense} &  89.20 \\ 
        DNN AVG + CKBG~\cite{saito-etal-2018-commonsense} & 94.70 \\ 
        Factorized~\cite{jastrzebski-etal-2018-commonsense} & 79.40 \\ 
        Prototypical~\cite{jastrzebski-etal-2018-commonsense} &  89.00 \\ 
        Concatenation~\cite{davison-etal-2019-commonsense} &  68.80 \\
        Template~\cite{davison-etal-2019-commonsense} &  72.20 \\ 
        Template + Grammar~\cite{davison-etal-2019-commonsense} &  74.40 \\ 
        Coherency Ranking~\cite{davison-etal-2019-commonsense} &  78.80 \\ 
        KG-BERT$_{\textrm{BERT-\textsc{Base}}}$~\cite{shen-etal-2020-exploiting} &  93.20 \\ 
        KG-BERT$_{\textrm{GLM(RoBERTa-\textsc{Large})}}$~\cite{shen-etal-2020-exploiting} & 94.60 \\ 
        \midrule 
        \textbf{Fine-tuned BERT (ours)} & \textbf{95.42} \\ 
        \textbf{Fine-tuned RoBERTa (ours)} & 94.37 \\ 
        \midrule 
        Human estimate~\cite{li-etal-2016-commonsense} & 95.00  \\ 
        \bottomrule
    \end{tabular}
    }
\end{table}

\section{Task-based evaluation}
\label{eval:extrinsic}

We next evaluate the efficacy of the ranking step in \negater{}.
Specifically, we next show how the top-ranking negative examples from \negater{} can be informative (\contradiction{}) for training \kb{} completion models. 
Similar to the previous section, we fine-tune pretrained BERT and RoBERTa models for a commonsense triple classification task. 
However, here we use a more challenging dataset split, and vary the ways that negatives are sampled at training time.

\subsection{Data}
As discussed previously, the \cn{} split introduced by~\citet{li-etal-2016-commonsense} is already saturated by BERT, likely because it contains ``easy'' negative test examples.
We therefore construct a new, more challenging split by taking the small percentage (3\pct{}) of triples in the benchmark with  negated relations (e.g., \rel{NotIsA}, six total), each of which has a positive counterpart in the \kb{} (e.g., \rel{IsA}). 
We filter the dataset to the positive/negated relation pairs only, and take the negated triples as \emph{true negative instances} for testing by removing the \rel{Not-} relation prefixes. 
Our new split, 
which we call \cn-TN to denote \underline{T}rue \underline{N}egatives, consists of 36,210/3,278/3,278 train/validation/test triples.
Again, the classes are balanced positive/negative, so accuracy is our main performance metric. 

Note that because this dataset contains true (hard) negatives, we expect accuracy to be much lower than what we achieved in Table~\ref{table:tc}.

\subsection{Baselines}

As baselines we consider several contrastive data augmentation approaches, all of which involve corrupting positive in-\kb{} samples.

We employ the following negative sampling baselines designed for \textbf{commonsense \kb{}s}:
\begin{itemize}
    \setlength{\itemsep}{0.0001em}
    \item \textbf{\uniform}~\cite{li-etal-2016-commonsense,saito-etal-2018-commonsense}: 
    We replace the head phrase $X_h$ or tail phrase $X_t$ of each positive $(X_h, r, X_t)^{+}$ by uniformly sampling another phrase from the \kb. 
    \item  \textbf{\comet}~\cite{bosselut-etal-2019-comet}: 
    \comet{} is a version of GPT~\cite{radford2018improving} that was fine-tuned to generate the tail phrase of a commonsense triple, conditioned on a head phrase and relation.
    To make \comet{} generate negatives, we prepend a ``not'' token to each positive head phrase $X_h^{+}$ and generate 10 tail phrases $X_{t}^{\textrm{\comet{}}}$ for the modified head/relation prefix using beam search.
    Finally, we replace the tail phrase $X_t$ in the positive with each $X_{t}^{\textrm{\comet{}}}$ in turn, yielding negatives $(X_h^{+}, r, X_{t}^{\textrm{\comet{}}})$. 
\end{itemize}
To investigate whether negative samplers tailored to encyclopedic knowledge can transfer to commonsense, we employ the following state-of-the-art baselines designed for \textbf{encyclopedic \kb{}s}:
\begin{itemize}
    \setlength{\itemsep}{0.0001em}
    \item \textbf{\selfadv}~\cite{sun-etal-2019-rotate}:  
    For each positive instance, a pool of candidate negatives is generated with \uniform{}. 
    The candidates are then scored by the (shallow, but state-of-the-art) RotatE \kb{} embedding, and a negative is sampled from the candidate pool with probability proportional to the score distribution.  
    We take the top 50\pct{} of self-adversarially generated statements as negative examples, in order of score descending, from the last epoch of training. 
    \item \textbf{\sans{}}~\cite{ahrabian-etal-2020-structure} is a graph-structural negative sampler that corrupts head/tail phrases of positive instances by sampling from the $k$-hop neighborhood of each \kb{} entity. 
    We set $k=2$.
\end{itemize}
Finally, we \textbf{devise two intuitive baselines}:
\begin{itemize} 
    \setlength{\itemsep}{0.0001em}
    \item \textbf{\slots}: We replace the head phrase $X_h$ (tail phrase $X_t$) of each positive $\triple^{+}$ by uniformly sampling from the set of phrases that appear in the head (tail) slot of \kb{} triples mentioning relation $r$. 
    We filter out all negative samples that appear in the \kb{}.
    \item \textbf{\antonyms}: We tag each phrase in the \kb{} as either a verb, noun, or adjective phrase using the SpaCy POS tagger.\footnote{\url{https://spacy.io/usage/linguistic-features\#pos-tagging}}
    Then, for each verb (noun, adjective) phrase, we replace the first verb (noun, adjective) token with a randomly selected antonym from either WordNet~\cite{miller1998wordnet} or the gold lexical contrast dataset from~\cite{nguyen-etal-2016-integrating}.
\end{itemize}

\subsection{\negater{} variants}
\begin{figure}
    \centering
    \includegraphics[width=0.7\columnwidth]{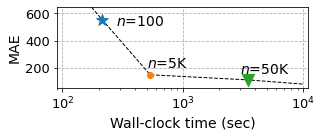}  
    \caption{
    Lower left corner is best: 
    Wall-clock time versus training loss (MAE) for \negaterGrad{} gradient magnitude prediction as training set size $n$ increases. 
    }
    \label{fig:grad-efficiency}
    \vspace{-.3cm}
\end{figure}

We generate out-of-\kb{} candidates for \negater{} with our $k$-NN approach using $k$=10, yielding around 570K candidates.
We implement the \negater{} candidate ranking methods as follows:
\begin{itemize}
    \setlength{\itemsep}{0.0001em}
    \item \textbf{\negaterThresh{}}: 
    We rank candidates using fine-tuned BERT's classification scores.
    Since the  scores are scaled differently by relation type, 
    we combine the top-ranking 50\pct{} of candidates per relation and shuffle them.
    \item \textbf{\negaterGrad{}}: 
    We again use BERT to rank the candidates.
    To choose between the full-gradient and gradient-prediction approaches (\S~\ref{method:negater-grad}),
    we train an MLP to predict gradient magnitudes and plot the mean absolute error training loss after 100 epochs for different training set sizes $n$. 
    Figure~\ref{fig:grad-efficiency} shows that even for $n$=5K examples, the loss quickly approaches zero. 
    Therefore, for efficiency, we use an MLP trained on $n$=20K examples, which takes around 1 hour to train and rank candidates on a single GPU, compared to an estimated 14 hours for the full-gradient approach.
    For a random sample of 100 candidates, the Pearson correlation coefficient between the true/predicted gradient magnitudes is $\rho$=0.982, indicating that the approximation is highly accurate.
    \item \textbf{No-ranking ablation}: Finally, in order to measure the importance of the \lm{} ranking component of \negater, we introduce an ablation which randomly shuffles the out-of-\kb{} candidates rather than ranking them. 
\end{itemize}
After we obtain each ranked list of candidates, we feed the statements as negative training examples to BERT/RoBERTa in order of rank descending. 

\subsection{Results}

For all performance metrics, we report averages over five trials to account for randomness in sampling and parameter initializations. 

\paragraph{Accuracy comparison}
As shown in Table~\ref{table:accuracy}, training with the top-ranking negative examples from \negater{} always yields the best accuracy for both \lm{}s, up to 1.90\pct{} more than the baselines.
Note that this improvement is achieved with changing how only \emph{half} of the training examples (the negatives) are sampled. 
Notice also that our \negater{} variants are the only samplers to offer \textbf{statistically significant improvements} over the \uniform{} baseline at $\alpha<0.01$ for BERT and $\alpha<0.05$ for RoBERTa (two-sided $t$-tests, five trials per model), signifying better-than-chance improvements. 

Notice also that our most competitive baseline is \slots{}, which is a contrastive approach that samples new head/tail phrases from those appearing in the corresponding slots of the current relation $r$---that is, preserving the grammar (\grammar) of the relation.
This confirms that grammatical negative samples are indeed more informative than nonsensical ones. 

\begin{table}[t!]
    \centering
	\caption{
	Accuracy on \cn{}-TN using different negative sampling approaches:
    Our \negater{} variants are the only negative samplers to offer statistically significant improvements over the popular \uniform{} baseline at $\alpha<0.01$ ($^{\blacktriangle}$) for BERT and $\alpha<0.05$ ($^{\triangle}$) for RoBERTa (two-sided $t$-test, five trials per model). 
	\underline{\textbf{Bold/underline}}: Best result per \lm; 
	\underline{Underline only}: Second-best result per \lm. 
	}
	\label{table:accuracy}
    \resizebox{0.99\columnwidth}{!}{
    \begin{tabular}{ c l ll }
        \toprule
        &  & BERT & RoBERTa \\ 
        \toprule 
        \multirow{6}{*}{\rot{90}{Baselines}} & \uniform{} & 75.60 $\pm$ 0.24 & 75.55 $\pm$ 0.43 \\  
        & \comet{} & 76.04 $\pm$ 0.63 & 75.86 $\pm$ 0.75 \\ 
        & \selfadv{} & 75.30 $\pm$ 0.51 & 75.20 $\pm$ 0.37 \\ 
        & \sans{} & 75.45 $\pm$ 0.38 & 75.17 $\pm$ 0.37 \\   
        & \slots{} & 76.46 $\pm$ 0.58$^\triangle$ & 75.80 $\pm$ 0.25 \\
        & \antonyms{} & 76.06 $\pm$ 0.30$^\triangle$ & 75.58 $\pm$ 0.58 \\
        \midrule 
        \multirow{3}{*}{\rot{90}{{\small \negater{}}}} & $\theta_r$ ranking & \textbf{\ul{76.95 $\pm$ 0.28}}$^{\blacktriangle}$ & \ul{76.29 $\pm$ 0.59} \\ 
        & $\nabla$ ranking & \ul{76.53 $\pm$ 0.22}$^{\blacktriangle}$ & \textbf{\ul{76.34 $\pm$ 0.32}}$^{\triangle}$ \\ 
        & No ranking & 75.61 $\pm$ 0.29 & 75.29 $\pm$ 0.19 \\
        \bottomrule
    \end{tabular}
    }
\end{table}

\paragraph{Encyclopedic versus commonsense?}
We hypothesize that the encyclopedic \kb{} baselines \selfadv{} and \sans{}  underperform because such methods assume that the \kb{} is a dense graph.
While this is usually true for encyclopedic \kb{}s,  
many commonsense \kb{}s are highly sparse because entities are not disambiguated, which means that multiple phrases referring to the same concept may be treated as different entities in the \kb.\footnote{\citet{malaviya-etal-2020-commonsense} provide an illustrative example for comparison: 
The popular encyclopedic \kb{} completion benchmark FB15K-237~\cite{toutanova-chen-2015-observed} is 75$\times$ denser than the \cn{} benchmark studied in this paper.}
For example, \sans{} assumes that there are plentiful entities within the $k$-hop neighborhood of a query entity, whereas in reality there may be very few, and these entities may not be grammatical in context of the original positive (\grammar) nor thematically relevant (\consistency) to the query entity. 
Therefore, encyclopedic negative samplers may not be transferrable to commonsense \kb{}s or other highly sparse \kb{}s. 

\paragraph{Ablation study}
Table~\ref{table:accuracy} also indicates that the \lm{} ranking component of \negater{} is crucial for improving accuracy. 
Our no-ranking ablation leads to lower classification accuracy than both \negaterThresh{} and \negaterGrad{}. 
Empirically, we find that this is because the ranking step helps filter out false negatives generated by our $k$-NN candidate construction procedure.

\paragraph{Performance drill-down}
Finally, Table~\ref{table:precision-recall} provides precision and recall scores to further ``drill down'' into \negater{}'s effects. 
Evidently, the \negater{} variants consistently yield the best precision, whereas there is no consistent winner in terms of recall.
To understand why \negater{} improves precision, we remind the reader that precision is calculated as $P = (TP) / (TP + FP)$, where $TP$ stands for true positives and $FP$ stands for false positives.
Because training with \negater{} examples helps the \lm{}s better recognize hard negatives---examples that ``look positive'' but are really negative---the \lm{} mislabels fewer negatives, decreasing the false positive rate. 

\begin{table}[t!]
    \centering
	\caption{
	\negater{} consistently yields the highest precision on \cn-TN  among negative samplers because it lowers the false positive rate:
	Performance drill-down (stdevs omitted for space). 
	$^{\blacktriangle}$, $^{\triangle}$: Significant improvement over \uniform{} at $\alpha < 0.01$ and $\alpha <0.05$, respectively. 
	}
	\label{table:precision-recall}
    \resizebox{0.99\columnwidth}{!}{
    \begin{tabular}{ c l ll c ll }
        \toprule
        & &  \multicolumn{2}{l}{BERT} & & \multicolumn{2}{l}{RoBERTa} \\ 
        \cline{3-4} \cline{6-7} 
        & & Prec. & Rec. && Prec. & Rec. \\ 
        \toprule 
        \multirow{6}{*}{\rot{90}{Baselines}} & \uniform{} &  71.29 & \textbf{\ul{85.83}}  && 73.36  & 80.28  \\  
        & \comet{} & 73.73  & 80.99  && 73.47  & \ul{81.02}  \\ 
        & \selfadv{} & 74.83 & 76.59   && 73.70  & 78.48  \\ 
        & \sans{} & 72.54  &  82.11  && 73.26  & 79.50  \\  
        & \slots{} & 75.21$^{\triangle}$  & 79.34  && 73.85  & 80.17  \\ 
        & \antonyms{} & 72.55  & \ul{83.98}  && 72.98  & \textbf{\ul{81.62}} \\ 
        \midrule 
        \multirow{3}{*}{\rot{90}{{\small \negater}}} & $\theta_r$ ranking & 75.12$^{\blacktriangle}$ & 80.68 && \textbf{\ul{75.92}}$^{\blacktriangle}$ & 77.05 \\
        & $\nabla$ ranking & \underline{76.60}$^{\blacktriangle}$ & 76.50 && \ul{75.75}$^{\blacktriangle}$ & 77.57 \\
        & No ranking & \textbf{\ul{76.81}}$^{\blacktriangle}$ & 73.42 && 75.67$^{\triangle}$ & 74.78 \\
        \bottomrule
    \end{tabular}
    }
\end{table}

\section{Human evaluation}
\label{eval:intrinsic}

Finally, we collect qualitative human judgments on the examples output by each negative sampler.

\subsection{Data}

To cover a diverse set of reasoning scenarios, we consider the \rel{HasPrerequisite}, \rel{HasProperty}, \rel{HasSubevent}, \rel{ReceivesAction}, and \rel{UsedFor} relations from \cn. 
For each relation and negative sampler, we take 30 negative statements at random, yielding 1,350 statements judged in total (5 relations $\times$ 9 negative samplers $\times$ 30 statements per method/relation).

\subsection{Methods}

We gather judgments for (\grammar{}) grammar on a binary scale (incorrect/correct) and (\consistency{}) thematic consistency of the head/tail phrases on a 4-point scale (``not consistent at all'', ``a little consistent'', ``somewhat consistent'', ``highly consistent'').
To estimate the true negative rate, we also obtain truthfulness judgments on a 4-point scale (``not truthful at all'', ``sometimes true'', ``mostly true'', ``always true''). 
We recruit four annotators who are fluent in English.
Among 50 statements shared across the annotators, we observe an average variance of 0.058 points on the 0/1 scale for \grammar{}, 0.418 points on the 4-point scale for \consistency{}, and 0.364 points on the 4-point truthfulness scale.
According to previous work in commonsense \kb{} construction~\cite{romero-etal-2019-commonsense}, these values indicate high agreement. 
Appendix~\ref{appendix:absolute-instructions} provides the annotation instructions.

\begin{table}[t!]
    \centering
    \renewcommand\thetable{5}
	\caption{
	\negater{} best trades off grammar (\grammar), consistency (\consistency), and the true negative rate, as measured by the percentage of statements labeled ``never true'':
	Human annotation scores, normalized out of 1. 
	Relative and average ranks are provided because not all raw metrics are directly comparable---e.g., grammar (\grammar) is judged as binary, whereas consistency (\consistency) is graded. 
	}
	\label{table:absolute-comparison}
    \resizebox{0.99\columnwidth}{!}{
    \begin{tabular}{ c l lll r  }
        \toprule
        &  & \grammar{} & \consistency{} & \pct{} ``never true'' & \cellcolor{gray!10} Avg rank \\ 
        \toprule
        \multirow{6}{*}{\rot{90}{Baselines}} & \uniform{} & 0.487 (9)  & 0.408 (7)  & 0.747 (3) & \cellcolor{gray!10} 6.33 (9) \\
        & \comet{} & 0.580 (8) & \textbf{\ul{0.703}} (1) & 0.407 (9) & \cellcolor{gray!10} 6.00 (8) \\
        & \selfadv{} & 0.733 (7)  & 0.373 (8) & \ul{0.767} (2) & \cellcolor{gray!10} 5.67 (7) \\
        & \sans{} &  0.760 (6)  & 0.532 (5)  & 0.633 (4) & \cellcolor{gray!10} 5.00 (4) \\
        & \slots{} & 0.853 (5) & 0.372 (9) & \textbf{\ul{0.773}} (1) & \cellcolor{gray!10} 5.00 (4) \\
        & \antonyms{} & 0.860 (4)  & 0.495 (6)  & 0.613 (5) & \cellcolor{gray!10} 5.00 (4) \\
        \midrule 
        \multirow{3}{*}{\rot{90}{{\small \negater}}} & $\theta_r$ ranking & 0.880 (3)  & \ul{0.635} (2) & 0.413  (8) & \cellcolor{gray!10} 4.33 (3) \\
        & $\nabla$ ranking & \textbf{\ul{0.927}} (1) & 0.555 (4) & 0.587 (6) & \cellcolor{gray!10} \textbf{\ul{3.67}} (1)  \\
        & No ranking & \ul{0.920} (2) & 0.592 (3)  & 0.560  (7) & \cellcolor{gray!10} \ul{4.00} (2) \\ 
        \bottomrule
    \end{tabular}
    }
\end{table}

\begin{table*}[t!]
\centering
\renewcommand\thetable{6}
\caption{
    Our \negaterGrad{} variant best handles the tradeoff between consistency (\consistency) and truthfulness: 
    Representative negative examples from the most competitive methods \slots{}, \negaterThresh{}, and \negaterGrad{}. 
}
\label{table:human-examples}
\resizebox{0.9\textwidth}{!}{
    \begin{tabular}{ l l r r c }
    \toprule
    Method & Negative statement & Consistent? & True? \\ 
    \toprule
    \multirow{3}{*}{\slots{}}   & (``open business'', \rel{HasPrerequisite}, ``hide behind door'') & A little &  Never \\ 
     & (``go somewhere'', \rel{HasSubevent}, ``bruise appears'') &  Not at all &  Never \\ 
     & (``mailbox'', \rel{UsedFor}, ``sleeping guests'') & Not at all &  Never \\ 
    \midrule 
    \multirow{3}{*}{\negaterThresh{}} & (``play baseball'', \rel{HasPrerequisite}, ``join hockey team'') & Somewhat & Never \\ 
     & (``comfort someone'', \rel{HasSubevent}, ``talk with them'') & Highly &  Mostly \\ 
     & (``having a bath'', \rel{UsedFor}, ``refreshing yourself'') & Highly & Sometimes \\ 
    \midrule 
    \multirow{3}{*}{\negaterGrad{}} & (``hear news'', \rel{HasPrerequisite}, ``record something'') & A little & Never \\ 
     & (``drink water'', \rel{HasSubevent}, ``inebriation'') & Highly & Never \\
     & (``luggage trolley'', \rel{UsedFor}, ``moving rocks'') & Highly & Never \\ 
    \bottomrule
\end{tabular}
}
\end{table*}

\subsection{Results}
Table~\ref{table:absolute-comparison} compares normalized average judgment scores for \grammar{} and \consistency{}, as well as the percentage of statements  labeled as ``never true'' (i.e., the true negative rate). 
Here, the takeaway is that the requirements of negative knowledge are a tradeoff, and \textbf{\negaterGrad{} best manages this tradeoff}. 
Indeed, the methods that yield the most true negatives (\selfadv{}, \slots{}) perform the worst in grammar (\grammar) and consistency (\consistency), whereas methods that yield more consistent statements like \comet{} have a comparatively low true negative rate. 

Finally, Table~\ref{table:human-examples} provides examples of statements with consistency  (\consistency) and truthfulness judgments. 
Again, it is evident that \negaterGrad{} best manages the tradeoffs of negative knowledge. 
In fact, it is the only negative sampler for which a majority of examples are rated both as ``never true'' (58.67\pct{}) and ``somewhat consistent'' or higher (62\pct{}).

\section{Related work}
\label{sec:related}
\paragraph{Commonsense \kb{} completion}
Approaches to commonsense \kb{} completion include  triple classification~\cite{li-etal-2016-commonsense,saito-etal-2018-commonsense,jastrzebski-etal-2018-commonsense,davison-etal-2019-commonsense}, generation of novel triples~\cite{bosselut-etal-2019-comet,hwang-etal-2021-comet-atomic},  and link prediction~\cite{malaviya-etal-2020-commonsense}. 
Such approaches either focus solely on modeling positive knowledge, or else generate negatives at random, making our work the first attempt in the direction of meaningful negative knowledge.

\paragraph{Knowledge in language models}
Several studies have shown that deep contextual language models acquire a degree of implicit commonsense knowledge during pretraining~\cite{petroni-etal-2019-language,davison-etal-2019-commonsense,roberts-etal-2020-much}, which can be further sharpened specifically toward \kb{} completion by targeted fine-tuning~\cite{bosselut-etal-2019-comet,hwang-etal-2021-comet-atomic}.
Our results in fine-tuning BERT to ConceptNet (\S~\ref{eval:lm}) validate these findings. 
Other studies have demonstrated that pretrained \lm{}s struggle to distinguish affirmative sentences from their negated counterparts~\cite{kassner-schutze-2020-negated,ettinger-2020-bert}, although this can again be addressed by fine-tuning~\cite{kassner-schutze-2020-negated,jiang-etal-2021-im}. 
Note, however, that we deal not with linguistic \emph{negation} but with negative, or false, knowledge. 

\paragraph{Negatives in knowledge bases}
We are not aware of any tailored negative samplers for commonsense knowledge.
However, several negative samplers for encyclopedic \kb{}s like Freebase exist, including self-adversarial~\cite{cai-wang-2018-kbgan,sun-etal-2019-rotate}, graph-structural~\cite{ahrabian-etal-2020-structure}, and heuristic ``interestingness''~\cite{arnaout-etal-2020-enriching} approaches. 
While these methods share our high-level goal, we showed in \S~\ref{eval:extrinsic} that they are less effective on highly sparse commonsense \kb{}s. 

\section{Conclusion}
\label{sec:conclusion}
In this paper we rigorously defined negative knowledge in commonsense \kb{}s and proposed a framework, \negater{}, to address this problem.
Importantly, our framework does not require ground-truth negatives at any point, making it an effective choice when gold training examples are not available. 
We empirically demonstrated the strength of \negater{} over many competitive baselines in multiple evaluations, including the strength of our fine-tuning approach, the task-based utility of \negater{} statements, and the intrinsic quality of these statements. 
A promising future direction is to explore new reasoning tasks that can be improved with negative knowledge from \negater{}, for example multiple-choice commonsense QA~\cite{ma-etal-2021-knowledge-driven}, which often requires high-quality negative examples for training.

\section*{Acknowledgements}
We thank the reviewers for their constructive feedback. 
This material is based upon work supported by the National Science Foundation under a Graduate Research Fellowship and CAREER Grant No.~IIS 1845491, Army Young Investigator Award No.~W911NF1810397, the Advanced Machine Learning Collaborative Grant from Procter \& Gamble, and Amazon, Google, and Facebook faculty awards.   
Any opinions, findings, and conclusions or recommendations expressed in this material are those of the author(s) and do not necessarily reflect the views of the National Science Foundation or other funding parties.

\balance
\bibliography{references}
\bibliographystyle{acl_natbib}

\newpage
\appendix
\section{Implementation details}
\label{appendix:fine-tuning} 

\paragraph{Pooling}
To obtain a single contextual representation of a triple from a sequence of triple tokens, we experiment with three standard pooling approaches~\cite{reimers-gurevych-2019-sentence}: Taking the reserved \CLS{} token embedding from the output of the encoder, and mean- and max-pooling over all output token representations.
As we do not observe statistically significant differences in performance among the pooling operations, we use the \CLS{} token as the triple embedding, since this is a very common method for encoding text sequences with BERT~\cite{gururangan-etal-2020-dont}. 

\paragraph{\lm{} hyperparameters}
Following the recommendations given by \citet{devlin-etal-2019-bert}, 
we search manually among the following hyperparameters (best configuration for BERT in \textbf{bold}, RoBERTa \ul{underlined}):
Batch size in $\{\underline{\mathbf{16}}, 32\}$; Learning rate in $\{10^{-4}, 10^{-5}, \underline{\mathbf{2 \times 10^{-5}}}, 3\times10^{-5}\}$; 
Number of epochs in $\{3, 5, \underline{7}, 10, \mathbf{13}\}$;
Number of warmup steps in $\{0, \underline{\mathbf{10\textrm{K}}}, 100\textrm{K}\}$;
Maximum sequence length in $\{16, \underline{\mathbf{32}}, 64\}$.
All other hyperparameters are as reported in~\cite{devlin-etal-2019-bert}.

\paragraph{Training negatives}
For the fine-tuning evaluation, to make our training process as similar as possible to common practice in the literature~\cite{li-etal-2016-commonsense,saito-etal-2018-commonsense,jastrzebski-etal-2018-commonsense}, we corrupt each positive $\xPos$  by replacing the head, relation, and tail of the positive in turn with a randomly sampled phrase or relation. 
For the task-based evaluation, we sample one negative per positive in order to make running many experiments across different negative samplers feasible.

\paragraph{Software and hardware}
We implement our \lm{}s with the Transformers PyTorch library~\cite{wolf-etal-2020-transformers} and run all experiments on a NVIDIA Tesla V100 GPU with 16 GB of RAM.
Both BERT and RoBERTa take around 1.5 hours/epoch to train on the \cn{} benchmark.

\section{Annotation instructions}
\label{appendix:absolute-instructions}

In this section we provide the annotation instructions for \S~\ref{eval:intrinsic}. 

\subsection{Task definition}

In this task you will judge a set of statements based on how grammatical, truthful, and consistent they are. Each statement is given in [head phrase, relation, tail phrase] form.
The criteria are as follows: 
\begin{itemize}
    \item \textbf{Grammar}: Our definition of grammar refers to whether each statement follows the grammar rules we provide for its relation type. We do not include proper use of punctuation (e.g., commas, apostrophes) or articles (e.g., ``the'', ``a'', ``this'') in our definition of grammar. The choices are ``correct'', ``partially correct or unsure'', and ``incorrect''. (\emph{Note: in our analyses we binarize these choices, considering ``partially correct'' and ``incorrect'' as the same.})
    \item \textbf{Truthfulness}: Our definition of truthfulness refers to how often you believe the whole statement holds true. The choices are: ``always true'', ``mostly true'', ``sometimes true'', and ``never true''.
    \item \textbf{Consistency}: We define ``consistency'' as the degree to which the head and tail phrases are consistent in terms of the topic, theme, or goal that they refer to. For example, the phrases ``football'' and ``baseball'' are highly consistent because they both refer to team sports, whereas the phrases ``football'' and ``cactus'' are not consistent. The choices are: ``highly consistent'', ``somewhat consistent'', ``a little consistent'', and ``not consistent at all''. 
\end{itemize}
You may fill your answers in any order. 
For example, you might find it helpful to judge the grammar of all statements first, then the truthfulness, then the consistency. 
Some of the statements are subjective and there is not always a ``right'' answer, especially for the consistency criterion. 
If you are unsure of a word or reference, you may use Google or other search engines.
You may also explain your reasoning/interpretation in the optional Notes box. 

\subsection{Examples and explanations}
\paragraph{HasPrerequisite}
The \rel{HasPrerequisite} relation describes prerequisites or pre-conditions for actions or states of being. It requires a verb phrase (an action or a state of being) in the head slot and a verb phrase or noun phrase in the tail slots. 
Examples:
\begin{itemize}
    \item (``pay bill'', \rel{HasPrerequisite}, ``have money'')
    \begin{itemize}
        \item Grammar: correct
        \item Truthfulness: always true
        \item Consistency: highly consistent
    \end{itemize}
    \item (``purchase a cellular phone'', \rel{HasPrerequisite}, ``study'')
    \begin{itemize}
        \item Grammar: correct
        \item Truthfulness: never true
        \item Consistency: not consistent at all
    \end{itemize}
    \item (``paint your house'', \rel{HasPrerequisite}, ``purple'')
    \begin{itemize}
        \item Grammar: incorrect
        \item Truthfulness: never true
        \item Consistency: a little consistent (\emph{Our interpretation: Painting your house involves choosing a color, so the statement could be construed as a little consistent, even though it's grammatically incorrect.})
    \end{itemize}
    \item (``eat'', \rel{HasPrerequisite}, ``send them to their room'')
    \begin{itemize}
        \item Grammar: correct
        \item Truthfulness: never true
        \item Consistency: not consistent at all
    \end{itemize}
\end{itemize}

\paragraph{HasProperty}
The \rel{HasProperty} relation describes properties of actions or objects. It requires a verb phrase or noun phrase in the head slot and a description in the tail slot. 
Examples:
\begin{itemize}
    \item (``school bus'', \rel{HasProperty}, ``yellow'')
    \begin{itemize}
        \item Grammar: correct
        \item Truthfulness: mostly true (\emph{Our interpretation: Yellow school buses are very common in the USA and Canada, but not all school buses are yellow.})
        \item Consistency: highly consistent
    \end{itemize}
    \item (``basketball'', \rel{HasProperty}, ``round'')
    \begin{itemize}
        \item Grammar: correct
        \item Truthfulness: always true
        \item Consistency: highly consistent
    \end{itemize}    
    \item (``pilot'', HasProperty, ``land airplane'')
    \begin{itemize}
        \item Grammar: incorrect
        \item Truthfulness: never true
        \item Consistency: highly consistent (\emph{Our interpretation: While pilots do land airplanes, the HasProperty relation requires a description in the tail slot, so it's not grammatically correct or truthful.})
    \end{itemize}
    \item (``gross domestic product'', \rel{HasProperty}, ``abbreviated to CTBT'')
    \begin{itemize}
        \item Grammar: correct
        \item Truthfulness: never true
        \item Consistency: a little consistent (\emph{Our interpretation: The gross domestic product does have a well-known abbreviation (``GDP''), so this statement could be construed as a little consistent.})
    \end{itemize}
\end{itemize}

\paragraph{HasSubevent}
The \rel{HasSubevent} relation describes sub-events or components of larger events. It requires an event (verb phrase or noun phrase) in the head slot and an event in the tail slot. 
Examples:
\begin{itemize}
    \item (``lying'', \rel{HasSubevent}, ``you feel guilty'')
    \begin{itemize}
        \item Grammar: correct
        \item Truthfulness: mostly true (\emph{Our interpretation: Lying often causes guilt in people, although the amount of guilt depends on the person.})
        \item Consistency: highly consistent
    \end{itemize} 
    \item (``relax'', \rel{HasSubevent}, ``vegetable'')
    \begin{itemize}
        \item Grammar: incorrect
        \item Truthfulness: never true
        \item Consistency: not consistent at all
    \end{itemize}
    \item (``drink coffee'', \rel{HasSubevent}, ``water may get into your nose'')
    \begin{itemize}
        \item Grammar: correct
        \item Truthfulness: never true
        \item Consistency: a little consistent (\emph{Our interpretation: Drinking coffee doesn't cause water to get into your nose, but coffee and water are both drinkable liquids, so we think this statement is a little consistent.})
    \end{itemize} 
\end{itemize}

\paragraph{ReceivesAction}
The \rel{ReceivesAction} relation describes actions that apply to objects or other actions. It requires a verb phrase or noun phrase in the head slot and an action in the tail slot. 
Examples:
\begin{itemize}
    \item (``book'', \rel{ReceivesAction}, ``write by person'')
    \begin{itemize}
        \item Grammar: correct
        \item Truthfulness: always true
        \item Consistency: highly consistent
    \end{itemize} 
    \item (``most watches'', \rel{ReceivesAction}, ``rhyme with piano'')
    \begin{itemize}
        \item Grammar: correct
        \item Truthfulness: never true
        \item Consistency: not consistent at all
    \end{itemize}
    \item (``oil'', \rel{ReceivesAction}, ``grow in field'')
    \begin{itemize}
        \item Grammar: correct
        \item Truthfulness: never true
        \item Consistency: a little consistent (\emph{Our interpretation: Since oil is a natural resource similar to other things that are grown in fields, we could see this statement being a little consistent (it's a stretch though)}.)
    \end{itemize} 
    \item (``violin'', \rel{ReceivesAction}, ``play with a puck'')
    \begin{itemize}
        \item Grammar: correct
        \item Truthfulness: never true
        \item Consistency: somewhat consistent (\emph{Our interpretation: Violins are indeed played, but with a bow, not a puck.})
    \end{itemize} 
\end{itemize}

\paragraph{UsedFor}
The \rel{UsedFor} relation describes the uses of objects or actions. It requires a verb phrase or noun phrase in the head and tail slots. 
Examples:
\begin{itemize}
    \item (``shoes'', \rel{UsedFor}, ``protecting feet'')
    \begin{itemize}
        \item Grammar: correct
        \item Truthfulness: always true
        \item Consistency: highly consistent
    \end{itemize}
    \item (``tying your shoelace'', \rel{UsedFor}, ``smart'')
    \begin{itemize}
        \item Grammar: incorrect
        \item Truthfulness: never true
        \item Consistency: not consistent at all
    \end{itemize}
    \item (``swimming'', \rel{UsedFor}, ``traveling on land'')
    \begin{itemize}
        \item Grammar: correct
        \item Truthfulness: never true
        \item Consistency: somewhat consistent (\emph{Our interpretation: This statement is somewhat consistent because swimming and traveling on land are both means of movement.})
    \end{itemize}
    \item (``bush'', \rel{UsedFor}, ``wrestling on'')
    \begin{itemize}
        \item Grammar: correct
        \item Truthfulness: never true
        \item Consistency: not consistent at all
    \end{itemize} 
\end{itemize}

\end{document}